\pdfoutput=1

\documentclass[11pt]{article}

\usepackage{ACL2023}

\usepackage{times}
\usepackage{latexsym}
\usepackage{booktabs}
\usepackage{amsmath}
\usepackage{graphicx}

\usepackage[T1]{fontenc}

\usepackage[utf8]{inputenc}

\usepackage{microtype}

\usepackage{inconsolata}

\title{UZH\_CLyp at SemEval-2023 Task 9: Head-First Fine-Tuning and ChatGPT Data Generation for Cross-Lingual Learning in Tweet Intimacy Prediction}

\author{Andrianos Michail \hspace{2em}  Stefanos Konstantinou \hspace{2em} {\bf Simon Clematide} \\
       Department of Computational Linguistics \\ University of Zurich \\ \texttt{\{andrianos.michail,stefanos.konstantinou,simon.clematide\}@uzh.ch}}
    
\begin{document}
\maketitle
\begin{abstract}
This paper describes the submission of UZH\_CLyp for the SemEval 2023 Task 9 ``Multilingual Tweet Intimacy Analysis''. We achieved second-best results in all 10 languages according to the official Pearson's correlation regression evaluation measure. Our cross-lingual transfer learning approach explores the benefits of using a Head-First Fine-Tuning method (HeFiT) that first updates only  the regression head parameters and then also updates the pre-trained transformer encoder parameters at a reduced learning rate. Additionally, we study the impact of using  a small set of automatically generated examples (in our case, from ChatGPT) for low-resource settings where no human-labeled data is available. Our study shows that HeFiT stabilizes training and consistently improves results for pre-trained models that lack domain adaptation to tweets. Our study also shows a noticeable performance increase in cross-lingual learning when  synthetic data is used, confirming the usefulness  of current text generation systems to improve zero-shot baseline results. Finally, we examine how possible inconsistencies in the  annotated data contribute to cross-lingual interference issues. 
\end{abstract}

\section{Introduction}

Social media texts are a rich source for studying social behavior, but manual analysis of them is prohibitively expensive. Predictive models that accurately estimate regression scores, such as intimacy, emotion, and valence, enable large-scale studies of social behavior through social media text.

Sparse training data in numerous languages presents another challenge to  globalize such studies of social behaviors. To overcome such limitations, this shared task focuses on developing models capable of accurately regressing intimacy in languages without training data.

In our shared task submission, we explore a novel approach to fine-tuning multilingual language models to maximize cross-lingual learning abilities. The key contributions and findings of this study are the following:

\begin{itemize}
\item We propose an alternative fine-tuning procedure for the XLM-R model that allows for an effective adaptation to the Twitter social media texts without further pre-training. Our results suggest that this approach yields slightly better performance in challenging cross-lingual learning of  unseen languages.
\item We demonstrate the effectiveness of using a small sample of synthetic tweets generated by  ChatGPT\footnote{\url{https://chat.openai.com/}} as training data for fine-tuning multilingual models. Our results suggest that this approach can effectively improve model performance in new languages.
\item We examine cross-lingual interference in our experimental framework and find that sometimes more training data of a language can surprisingly impair the performance of the language and related languages in the dataset.

\end{itemize}

\section{Related Work}

Cross-lingual learning involves the transfer of knowledge between natural languages, which can be achieved through the use of multilingual language models (MLMs). MLMs are pre-trained in multiple languages and then fine-tuned using available training data, with the aim of achieving positive transfer to languages with limited or no training data. One such MLM is XLM-R \citep{conneau2020unsupervised}, an XLM model \citep{lample2019cross} pre-trained with RoBERTa objectives \citep{liu2019roberta}, which has shown good multilingual performance. Another model, XLM-T, extends the XLM-R model with additional pre-training on Tweets, enhancing its performance on multilingual and monolingual tasks \citep{barbieri-etal-2022-xlm}.

Recent advances in multilingual models aim to enhance their adaptation to new languages by using of lightweight language adapter parameters \citep{pfeiffer2020mad}. Subsequent research has aimed to improve the performance of these models in cross-lingual transfer tasks by implementing language adapters whose weights are generated by Contextual Parameter Generation networks \citep{platanios2018contextual,ansell2021mad}. This approach has also been extended to handle multilingual multitask learning scenarios \citep{ustun2022hyper}.

Our hypothesis is that the adaptor-free architecture of MLMs is adequate to support cross-lingual learning when accompanied by our proposed staged fine-tuning and synthetic data generation pipeline that allows the models to improve performance in an unseen language with minimal compromise of performance for the training languages.

\section{Material and Methods}

\subsection{Dataset}

The dataset, which has been prepared by the organizers, consists of a collection of tweets that are accompanied by intimacy values as defined in \citet{pei2020quantifying}. The tweets are annotated on a 5-point Likert scale where a score of 1 indicates ''Not intimate at all" and a score of 5 indicates ''Very intimate". More information on the dataset and annotation process is available in the task paper \citep{Pei-EtAl:2023:SemEval}.
The dataset contains approximately 1,600 samples for each of the five Indo-European languages and Chinese. The test dataset includes ten languages, namely the six  languages from the training set plus Dutch, Arabic, Korean, and Hindi. Detailed information about the data sets can be found in Table~\ref{tab:dataset_stats}.

\begin{table}[t] 
\centering
\resizebox{\linewidth}{!}{%

\begin{tabular}{ccccc}
\toprule
\textbf{Language}  & \textbf{Training Set} & \textbf{Dev Set} & \textbf{Synth Set} & \textbf{Test Set} \\
\midrule
English  & 1,270 & 317 & 50 & 396 \\
Spanish  & 1,274 & 318 & 50 & 399 \\
Portuguese & 1,285 & 311 & 50 & 398 \\
Italian & 1,226 & 306 & 50 & 384 \\
French & 1,271 & 317 & 50 & 393\\
Chinese  & 1,277 & 319 & 50 & 400\\
\midrule
Hindi  & 0 & 0 & 50 & 280 \\
Korean  & 0 & 0 & 50 & 411 \\
Dutch  & 0 & 0 & 50 & 413 \\
Arabic  & 0 & 0 & 50 & 407 \\
\midrule
Total &  7,603  &  1,888  &  500 & 3,881\\
\bottomrule
\end{tabular}

}
\caption{Sizes of all data sets.}
\label{tab:dataset_stats}

\end{table}

\subsection{ChatGPT-Generated Data Samples}

To examine the ability of the multilingual model to quickly adapt to a new language, we generated a synthetic dataset consisting of fifty labeled samples for all ten test set languages. The model that generates these samples is the ChatGPT research preview, released on the 15th of December. Preliminary studies \citep{jiao2023chatgpt,bang2023multitask} suggest that ChatGPT has a competitive performance against established commercial translation systems, demonstrating its ability to generate multilingual content. However, the specific structure and components of the ChatGPT model have not been disclosed at the time of writing.

The textual prompt was designed based on empirical evidence, and subsequent modifications were made accordingly. The specific textual prompt used in the study is shown in Appendix  \ref{app:appendix_prompt}.

Although post-submission experiments showed minimal benefit (as demonstrated in Table~\ref{tab:annotation}), a simple validation process was implemented to ensure the data is at least of some quality. This distant pseudo-annotation process involved presenting 10 batches of 10 labeled items to a native speaker of the language, who selected the 5 best batches based on a provided definition of the intimacy score and 20 English reference samples \cite{pei2020quantifying}. Annotators were also warned that offensive or sexual content might be present in the samples or generated data. 

The annotation process for each language involved the participation of a single native speaker. The generation of this  dataset required a total workload of approximately 10 hours. The resulting synthetic dataset, which comprises a total of 500 data samples,  will be made publicly available.

In this study, the term `few-shot experiments' refers to the traditional definition of few-shot learning, where the model is instead exposed to a small subset of the training data points (specifically 50) of a specific language during the fine-tuning phase, while the inference procedure remains unchanged. Concerning the synthetic few-shot experiments, all 50 data samples are synthetic.

\begin{table*}[t]
\resizebox{\linewidth}{!}{%

\begin{tabular}{lccccccc}
\hline Model & Overall & English & Spanish & Italian & Portuguese & French & Chinese  \\
\hline 
XLM - RoBERTa $_{base}$ & & & & & & \\
$\quad$ SFiT (Hyp Set 1) & $0.669$ & $0.706$ & $0.604$ & $0.620$ & $0.592$ & $0.673$ & $0.761$ \\
$\quad$ SFiT (Hyp Set 2) & $0.662$ & $0.701$ & $0.617$ & $0.610$ & $0.578$ & $0.655$ & $0.750$ \\
$\quad$ HeFiT & $\mathbf{0.685}$ & $\mathbf{0.724}$ & $\mathbf{0.617}$ & $\mathbf{0.628}$ & $\mathbf{0.637}$ & $\mathbf{0.685}$ & $\mathbf{0.767}$ \\
XLM - Tweet  RoBERTa $_{base}$ & & & & & & \\
$\quad$ SFiT (Hyp Set 1) & $0.707$ & $0.751$ & $0.672$ & $\mathbf{0.652}$ & $0.659$ & $0.700$ & $0.761$ \\
$\quad$ SFiT (Hyp Set 2)  & $0.708$ & $0.755$ & $\mathbf{0.678}$ & $0.640$ & $0.658$ & $0.703$ & $0.761$ \\
$\quad$ HeFiT  & $\mathbf{0.710}$ & $\mathbf{0.756}$ & $0.669$ & $0.638$ & $\mathbf{0.669}$ & $\mathbf{0.711}$ & $\mathbf{0.764}$ \\
\hline
\end{tabular}
}
\caption{Pearson Correlation on our internal validation set. Hyp Set refers to the best set of hyperparameters found for the specific architecture, as detailed in Appendix \ref{app:hyp_space}. Results are averaged over 10 successful training runs.}
\label{tab:val_set_results}
\end{table*}

\subsection{Evaluation Metric}

The official evaluation of the shared task is the Pearson correlation coefficient score (Pearson's r).
This metric measures the linear correlation between two data sets. It is a normalized quantification of covariance, and the resulting values are  between -1 and 1. The greater the distance from zero, the stronger the relationship between the predictions and the true scores.

The Pearson correlation coefficient is defined by the following formula:
\[ \scalebox{1}{$
    r_{xy}=\frac{\sum_{i=1}^n\left(x_i-\bar{x}\right)\left(y_i-\bar{y}\right)}{\sqrt{\sum_{i=1}^n\left(x_i-\bar{x}\right)^2} \sqrt{\sum_{i=1}^n\left(y_i-\bar{y}\right)^2}}
$} \]

Where $y$ are the predictions of the models and $x$ are the labels of the validation/test set.

Our paper presents results in Pearson's r for both language-specific data evaluation and evaluation over all available languages. 

We acknowledge limitations in evaluating using this metric, as both the numerator and denominator rely on the average scores of the predictions, which can be strategically post-processed to improve evaluation.

\subsection{Problem Modelling}
We limit our investigation to evaluating the performance of fine-tuning XLM-T \citep{barbieri-etal-2022-xlm} and XLM-R base variants \citep{conneau2020unsupervised}. We make use of these pre-trained multilingual transformers encoders by attaching a regression head on the language model. Subsequently, we fine-tune the whole model using the training data.

Our study aims to make the findings applicable to additional regression tasks. Hence, we fine-tuned all models using the Mean Squared Error (MSE) loss function, a commonly used regression loss function. It is possible that our models could be at a disadvantage compared to models trained using a differentiable Pearson's r since we have fine-tuned them with the MSE loss function.

\subsection{Head-First Fine-tuning}
Low amounts of labeled data and short sample lengths (e.g., tweets) can cause unstable Encoder Transformer model fine-tuning  in regression tasks \citep{howard2018universal}. To address this issue, we propose an alternative fine-tuning approach inspired by the concept of gradual unfreezing as introduced in \citet{howard2018universal}.

Head-First Fine-Tuning (HeFiT) is a two-step procedure to fine-tune Encoder Transformers used for regression. We add  dropout to the hidden layer of the head. In the first step the embedding layers are frozen and the regression head is fine-tuned for three epochs.  In the second step, the entire model is unfrozen and fine-tuned  for six epochs with the learning rate halved.  The rationale behind the approach  is that the random weight initialization of the regression head in the regular fine-tuning may lead to  inaccurate signals being propagated back to the embedding layers, resulting in degradation of the pre-trained model. However, our investigation extends beyond HeFiT. Our experimental design incorporates an exploration of the regular fine-tuning procedure, which we shall refer to as Standard Fine-Tuning (SFiT) for ease of reference. The concept known as HP-FT, closely related to HeFiT, has also been recently investigated in the field of computer vision \citep{renprepare,kumar2022fine}. Our initial experiments indicate that this approach brings greater improvements when there is a discrepancy between the pre-training and the fine-tuning text domain and possibly also in the cross-lingual learning scenario.

\begin{figure*}[t]
\centering
\includegraphics[width=\linewidth]
{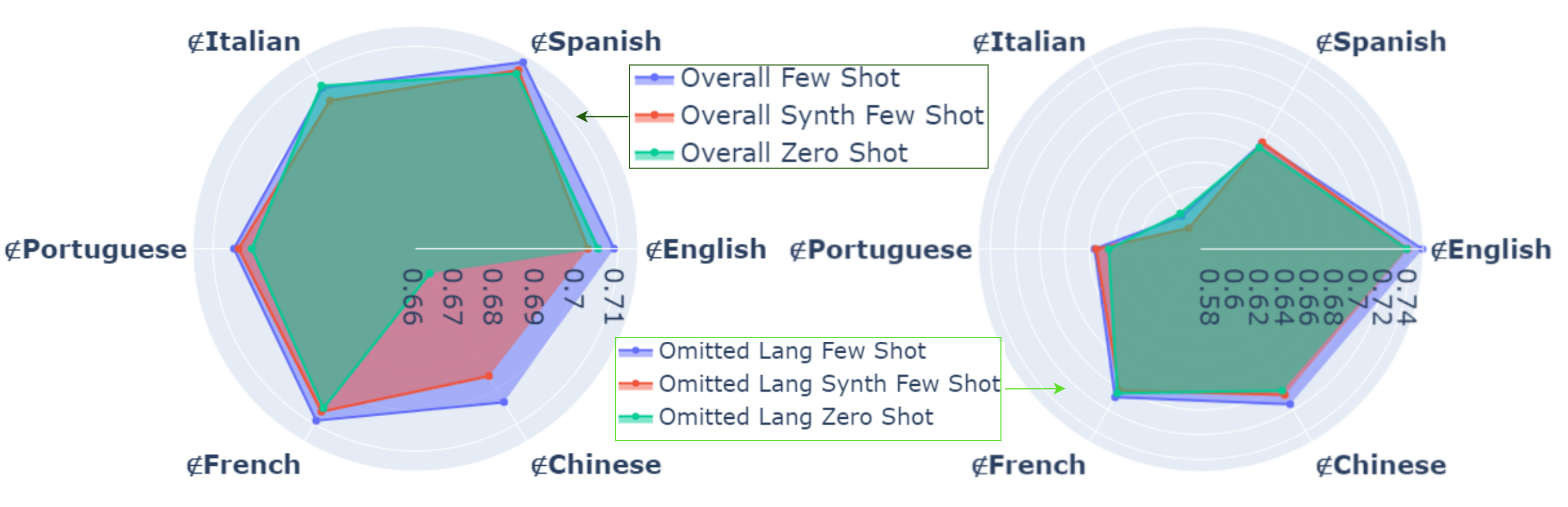}
\caption{Radar plot of Pearson's r of the  HeFiT fine-tuned XLM-T model on the complete dev set, calculated as an average over 10 runs. The left plot shows the overall performance and the right plot shows the performance of the omitted language. The results of SFit and XLM-R  are  analogous (not shown).}
\label{fig:radar_few}
\end{figure*}

\subsection{Final Submission Ensembling}

Experimental results on the validation set demonstrate improved stability and performance by combining predictions (i.e. average ensemble) of predictions of models trained using different fine-tuning methods, consistent with \citet{michail2021uzh}. Our final predictions are an ensemble of six XLM-T HeFiT models and four SFit fine-tuned models trained on different subsets of the available data. Further details on the specific models used for our submission can be found in Appendix \ref{tab:post_sub_models}.

\section{Results and Discussion}

\subsection{Performance Ablation Study}

In an exploratory investigation, a grid search was used to determine two satisfactory configurations of fine-tuning hyperparameters to optimize the performance of the validation set. In our further experiments, we kept the same hyperparameters. Information about the explored and chosen hyperparameters can be found in Appendix \ref{app:hyp_space}.

As shown in Table~\ref{tab:val_set_results}, HeFiT fine-tuning, when paired with the XLM-R model, exhibits a notable competitive advantage in the task at hand. In contrast, this advantage is less apparent in the XLM-T model. Based on this observation, we posit that the HeFiT model enables XLM-R to shift more seamlessly towards the Twitter domain, while XLM-T does not need to, resulting in minimal performance gains.

\subsection{Synthetic Data Samples Performance Analysis}

In order to assess the efficacy of synthetic data samples in enhancing the performance of cross-lingual learning, we conduct simulations with our training languages. Specifically, we systematically omit the training data of a single language (Zero Shot) and report the overall performance of all languages on the validation set. Furthermore, we assess the performance of the same configuration in scenarios where only 50 samples of the omitted language (Few Shot) or 50 synthetic samples (Synth Few Shot) are available. Figure~\ref{fig:radar_few} illustrates the overall performance and the performance of the omitted language  for these experiments.

Beginning with the negative results, the addition of 50 samples, particularly synthetic ones, results in a decrease in performance for Italian. This may be due to peculiarities in the training and validation data for this language.

Conversely, the remaining Romance languages show a comparable improvement in both overall and language-specific performance metrics when  either training or synthetic training samples are used.

An additional noteworthy finding is that the language in our training dataset exhibiting the highest degree of deviation, namely Chinese, experiences a significant performance degradation in the Zero Shot scenario. This is evidenced by  comparison with the  results of the experiments presented in Table~\ref{tab:val_set_results}. It is noteworthy that significant performance improvements are observed with the inclusion of 50 Chinese training samples, particularly in the overall performance metric, which can be attributed to increased prediction stability. A similar effect is discernible, albeit to a lesser degree, when employing synthetic training samples.

\subsection{Limiting Training Data for Interfering Languages}

We found that omitting Italian or Spanish training data noticeably improved the performance of the remaining Romance languages. To further examine this effect, we conducted experiments with various training sample sizes for the controlled languages, Italian and Spanish.

\begin{figure}[t]
\centering
\includegraphics[width=\linewidth]
{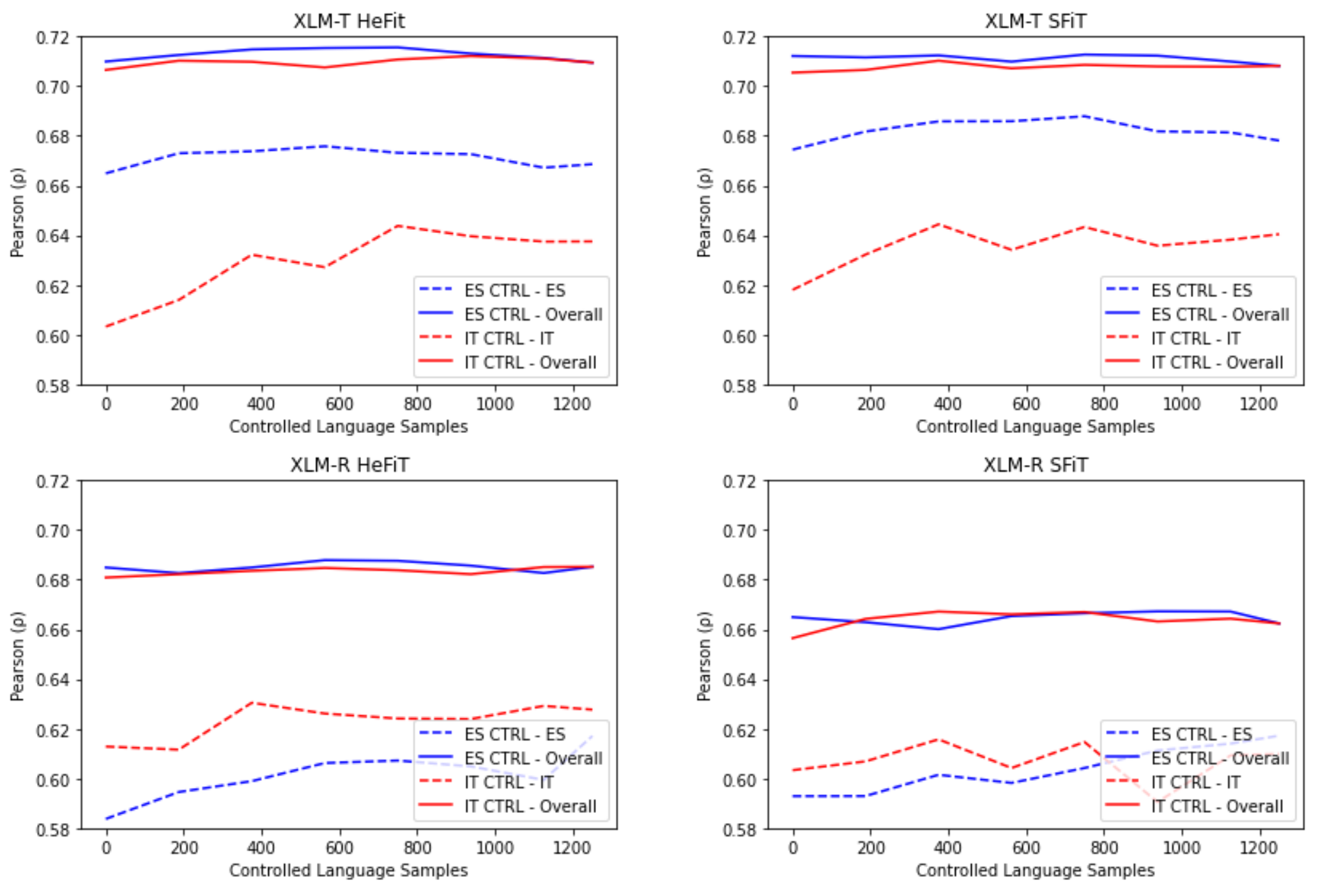}
\caption{Pearson's r of the validation set, calculated as the mean value across 10 successful training iterations, while controlling the sample size for the Spanish (ES) and Italian (IT) datasets.}
\label{fig:latin_abl}
\end{figure}

Figure \ref{fig:latin_abl} suggests that the ideal number of samples for optimal model performance whilst maintaining controlled language performance is between 550-800 per language. An interesting observation is that increased training data for a controlled language can lead to interference with performance in related languages, including within the controlled language (e.g., Spanish in XLM-T). However, sample quality and labeling consistency across languages are critical factors in this effect. Hence, we postulate that this phenomenon might be limited to this specific dataset, and its generalizability to other datasets remains uncertain.

\subsection{Post Submission Evaluations}

To evaluate cross-lingual performance, we perform experiments on the test set as reported in Table~\ref{tab:post_sub_results}.
In all XLM-R results, HeFiT significantly improves performance, supporting the hypothesis that it enables effective domain adaptation.

For XLM-T, SFiT demonstrates better performance in the zero-shot scenario. When considering aggregated results of models that were finetuned with some synthetic data, the performance difference between the methods becomes negligible. However, when the synthetic data is limited to Korean and Hindi, HeFiT improves performance. Details about the models evaluated can be found in Table~\ref{tab:post_sub_models}.

A general observation is that synthetic training data consistently improves performance, resulting in overall better multilingual prediction models.

\subsection{Experimental Setup}
To fine-tune the models, we used MSELoss and utilized the Simple Transformers \footnote{\url{https://simpletransformers.ai}} library, a wrapper for the Hugging Face \footnote{\url{https://huggingface.co}} library, that allows for fast experimenting. All models were trained on a single T4 GPU instance.
\begin{table}[t]
\resizebox{\linewidth}{!}{%

\begin{tabular}{lcccc}
\hline & All Models & Synth FewS & ZeroS & Synth\{KO,HI\}\\
\hline 
XLM-R & & & &  \\
$\quad$ SFiT & $0.548$ & $0.549$ & $0.541$ & $0.556$ \\
$\quad$ HeFiT & $\mathbf{0.558}$ & $\mathbf{0.560}$ & $\mathbf{0.548}$ & $\mathbf{0.562}$ \\
XLM-T & & & & \\
$\quad$ SFiT & $0.602$ & $0.604$ & $\mathbf{0.599}$ & $0.603$ \\
$\quad$ HeFiT & $\mathbf{0.603}$ & $\mathbf{0.605}$ & $0.589$ & $\mathbf{0.610}$ \\
\hline
\end{tabular}
}
\caption{Pearson's r for all test set items reported for each pre-trained model and fine-tuning technique. The results are averages from our corresponding 15 training data compilation settings, which leverage distinct subsets of (synthetic) training data as detailed in Table~\ref{tab:post_sub_models}.}
\label{tab:post_sub_results}

\end{table}

\section{Conclusion}

This system description paper presents and discusses  our submission, which was ranked second and achieved a Pearson's r of 0.614 in the Multilingual Twitter Intimacy Regression task of SemEval 2023. We propose a fine-tuning approach that improves XLM-R's adaptation to Twitter and demonstrates minor improvements in cross-lingual learning. Additionally, we generate a mini-dataset of synthetic tweets in all 10 task languages using ChatGPT, which suffers only minor performance loss in a few-shot scenario against manually annotated training data. Finally, we observe indications of self and cross-lingual interference for Romance languages in both models, and we would like to investigate if this phenomenon is replicable in other tasks and datasets.

\section*{Limitations}
The fine-tuning task investigated in this paper is limited to the tweet intimacy regression task, and our findings may not be replicated in other datasets or tasks, even within the same domain. Therefore, future research is needed to confirm the effects of these techniques in a variety of tasks and domains. Another limitation is our subjective design of a text prompt to generate synthetic training data, which may not be the best possible prompt. In addition, we used ChatGPT for multilingual synthetic data generation, and its internal mechanisms are undisclosed at the time of writing.



\section*{Acknowledgements}
We want to thank the Department of Computational Linguistics for providing us with the technical infrastructure.

\bibliography{anthology,custom}

\begin{thebibliography}{16}
\expandafter\ifx\csname natexlab\endcsname\relax\def\natexlab#1{#1}\fi

\bibitem[{Ansell et~al.(2021)Ansell, Ponti, Pfeiffer, Ruder, Glava{\v{s}},
  Vuli{\'c}, and Korhonen}]{ansell2021mad}
Alan Ansell, Edoardo~Maria Ponti, Jonas Pfeiffer, Sebastian Ruder, Goran
  Glava{\v{s}}, Ivan Vuli{\'c}, and Anna Korhonen. 2021.
\newblock {MAD-G}: Multilingual adapter generation for efficient cross-lingual
  transfer.
\newblock In \emph{Findings of the Association for Computational Linguistics:
  EMNLP 2021}, pages 4762--4781.

\bibitem[{Bang et~al.(2023)Bang, Cahyawijaya, Lee, Dai, Su, Wilie, Lovenia, Ji,
  Yu, Chung et~al.}]{bang2023multitask}
Yejin Bang, Samuel Cahyawijaya, Nayeon Lee, Wenliang Dai, Dan Su, Bryan Wilie,
  Holy Lovenia, Ziwei Ji, Tiezheng Yu, Willy Chung, et~al. 2023.
\newblock A multitask, multilingual, multimodal evaluation of {ChatGPT} on
  reasoning, hallucination, and interactivity.
\newblock \emph{arXiv preprint arXiv:2302.04023}.

\bibitem[{Barbieri et~al.(2022)Barbieri, Espinosa~Anke, and
  Camacho-Collados}]{barbieri-etal-2022-xlm}
Francesco Barbieri, Luis Espinosa~Anke, and Jose Camacho-Collados. 2022.
\newblock \href {https://aclanthology.org/2022.lrec-1.27} {{XLM}-{T}:
  Multilingual language models in {T}witter for sentiment analysis and beyond}.
\newblock In \emph{Proceedings of the Thirteenth Language Resources and
  Evaluation Conference}, pages 258--266, Marseille, France. European Language
  Resources Association.

\bibitem[{Conneau et~al.(2020)Conneau, Khandelwal, Goyal, Chaudhary, Wenzek,
  Guzm{\'a}n, Grave, Ott, Zettlemoyer, and Stoyanov}]{conneau2020unsupervised}
Alexis Conneau, Kartikay Khandelwal, Naman Goyal, Vishrav Chaudhary, Guillaume
  Wenzek, Francisco Guzm{\'a}n, {\'E}douard Grave, Myle Ott, Luke Zettlemoyer,
  and Veselin Stoyanov. 2020.
\newblock Unsupervised cross-lingual representation learning at scale.
\newblock In \emph{Proceedings of the 58th Annual Meeting of the Association
  for Computational Linguistics}, pages 8440--8451.

\bibitem[{Howard and Ruder(2018)}]{howard2018universal}
Jeremy Howard and Sebastian Ruder. 2018.
\newblock Universal language model fine-tuning for text classification.
\newblock In \emph{Proceedings of the 56th Annual Meeting of the Association
  for Computational Linguistics (Volume 1: Long Papers)}, pages 328--339.

\bibitem[{Jiao et~al.(2023)Jiao, Wang, Huang, Wang, and Tu}]{jiao2023chatgpt}
Wenxiang Jiao, Wenxuan Wang, Jen-tse Huang, Xing Wang, and Zhaopeng Tu. 2023.
\newblock Is {ChatGPT} a good translator? a preliminary study.
\newblock \emph{arXiv preprint arXiv:2301.08745}.

\bibitem[{Kumar et~al.(2022)Kumar, Raghunathan, Jones, Ma, and
  Liang}]{kumar2022fine}
Ananya Kumar, Aditi Raghunathan, Robbie Jones, Tengyu Ma, and Percy Liang.
  2022.
\newblock Fine-tuning can distort pretrained features and underperform
  out-of-distribution.
\newblock In \emph{International Conference on Learning Representations}.

\bibitem[{Lample and Conneau(2019)}]{lample2019cross}
Guillaume Lample and Alexis Conneau. 2019.
\newblock Cross-lingual language model pretraining.
\newblock \emph{arXiv preprint arXiv:1901.07291}.

\bibitem[{Liu et~al.(2019)Liu, Ott, Goyal, Du, Joshi, Chen, Levy, Lewis,
  Zettlemoyer, and Stoyanov}]{liu2019roberta}
Yinhan Liu, Myle Ott, Naman Goyal, Jingfei Du, Mandar Joshi, Danqi Chen, Omer
  Levy, Mike Lewis, Luke Zettlemoyer, and Veselin Stoyanov. 2019.
\newblock {Roberta}: A robustly optimized {BERT} pretraining approach.
\newblock \emph{arXiv preprint arXiv:1907.11692}.

\bibitem[{Michail et~al.(2021)Michail, Wehrli, and
  Buckov{\'a}}]{michail2021uzh}
Adrianos Michail, Silvan Wehrli, and Ter{\'e}zia Buckov{\'a}. 2021.
\newblock {UZH OnPoint at Swisstext-2021}: Sentence end and punctuation
  prediction in {NLG} text through ensembling of different transformers (short
  paper).
\newblock In \emph{SwissText}.

\bibitem[{Pei and Jurgens(2020)}]{pei2020quantifying}
Jiaxin Pei and David Jurgens. 2020.
\newblock Quantifying intimacy in language.
\newblock In \emph{Proceedings of the 2020 Conference on Empirical Methods in
  Natural Language Processing (EMNLP)}.

\bibitem[{Pei et~al.(2023)Pei, Silva, Bos, Liu, Neves, Jurgens, and
  Barbieri}]{Pei-EtAl:2023:SemEval}
Jiaxin Pei, V{\'\i}tor Silva, Maarten Bos, Yozon Liu, Leonardo Neves, David
  Jurgens, and Francesco Barbieri. 2023.
\newblock Semeval 2023 task 9: Multilingual tweet intimacy analysis.
\newblock In \emph{Proceedings of the 17th International Workshop on Semantic
  Evaluation}, Toronto, Canada. Association for Computational Linguistics.

\bibitem[{Pfeiffer et~al.(2020)Pfeiffer, Vuli{\'c}, Gurevych, and
  Ruder}]{pfeiffer2020mad}
Jonas Pfeiffer, Ivan Vuli{\'c}, Iryna Gurevych, and Sebastian Ruder. 2020.
\newblock Mad-x: An adapter-based framework for multi-task cross-lingual
  transfer.
\newblock In \emph{Proceedings of the 2020 Conference on Empirical Methods in
  Natural Language Processing (EMNLP)}, pages 7654--7673.

\bibitem[{Platanios et~al.(2018)Platanios, Sachan, Neubig, and
  Mitchell}]{platanios2018contextual}
Emmanouil~Antonios Platanios, Mrinmaya Sachan, Graham Neubig, and Tom Mitchell.
  2018.
\newblock Contextual parameter generation for universal neural machine
  translation.
\newblock In \emph{Proceedings of the 2018 Conference on Empirical Methods in
  Natural Language Processing}, pages 425--435.

\bibitem[{Ren et~al.(2023)Ren, Guo, Bae, and Sutherland}]{renprepare}
Yi~Ren, Shangmin Guo, Wonho Bae, and Danica~J Sutherland. 2023.
\newblock How to prepare your task head for finetuning.
\newblock In \emph{The Eleventh International Conference on Learning
  Representations}.

\bibitem[{{\"U}st{\"u}n et~al.(2022){\"U}st{\"u}n, Bisazza, Bouma, van Noord,
  and Ruder}]{ustun2022hyper}
Ahmet {\"U}st{\"u}n, Arianna Bisazza, Gosse Bouma, Gertjan van Noord, and
  Sebastian Ruder. 2022.
\newblock \href {https://aclanthology.org/2022.emnlp-main.541} {Hyper-{X}: A
  unified hypernetwork for multi-task multilingual transfer}.
\newblock In \emph{Proceedings of the 2022 Conference on Empirical Methods in
  Natural Language Processing}, pages 7934--7949, Abu Dhabi, United Arab
  Emirates. Association for Computational Linguistics.

\end{thebibliography}
\bibliographystyle{acl_natbib}

\newpage

\appendix
\onecolumn
\section{Textual Prompt for ChatGPT}
\label{app:appendix_prompt}

Warning: The prompt contains data from the training set, hence you might see offensive or sexual content in this appendix.\\

\fbox{
\begin{minipage}{\linewidth}
Type of Text: Short Tweets (Maximum of 10 words). 
The tweets may include errors slang, emojis, mentions and hashtags. \\

Formality: Informal \\

Intimacy criterion: Intimacy is a fundamental aspect of how we relate to others in social settings. Language encodes the social information of intimacy through the privacy of the topics and other cues such as linguistic hedging and very importantly swearing.\\
\\
Intimacy distribution statistics:\\
Mean: 2.1\\
Minimum: 1\\
Maximum: 5\\
Standard Deviation: 0.9\\
\\
Examples:\\
@user And then you change into your NSFW account and like them 	Intimacy Score: 2.6/5\\
But I, I can feel it take a hold I vote \#WatermelonSugar as \#BestMusicVideo at the \#iHeartAwards	Intimacy Score: 1/5\\
Who should I draw on my live to entertain the horny mfs I know are gonna show up	Intimacy Score: 3/5\\
@user @mehdirhasan in addition, how can you change a rule written by god. he didn't just change his mind	Intimacy Score: 2.2/5\\
@user Holy crap! Suzi took that photo! That’s hilarious!	Intimacy Score: 2.8/5\\
Dragon angling darma	Intimacy Score: 2.25/5\\
like seriously just say ur pretending to like her cause of ariana and leave	Intimacy Score: 1.75/5\\
@user I think I fell in love with you	Intimacy Score: 4.8/5\\
@user sis didn't you get your nipples pierced??? ur the bravest woman alive already. whats a lil ear needle 	Intimacy Score: 4.4/5\\
I need you mf to step your game up cause the way this man came at me I almost threw up .	Intimacy Score: 2.33/5\\
@user @user This is how it would be for me. It's humiliating, and embarrassing, and a waste of my time and energy...	Intimacy Score: 3.2/5\\
@user Beyonc\'e is overrated	Intimacy Score: 1.2/5\\
lowkey wanna talk to him just like the old days	Intimacy Score: 2.5/5\\
\\
Generate 10 more Tweet samples in [TGT language] including their Intimacy Score according to the Intimacy criterion:

\end{minipage}
}
\twocolumn

\section{Hyperparameter Space}
For Standard FineTuning, the batch size was fixed at 8 and the learning rate at $4\mathrm{e}{-5}$. For each model, we examined 4 to 10 epochs and the following  dropout values of the hidden layer: $\{0.00, 0.05, 0.1, 0.15, 0.2\}$.

For Head-First Fine-Tuning, we selected a number of hyperparameters from previous work on a regression task and similarly examined their performance on this task. Empirically, we chose 0.05 hidden dropout probability. We believe that the best hyperparameters and procedure details for Head-First Fine-Tuning remain unexplored.

For the performance ablation study, we selected the two best sets of hyperparameters for SFiT for a double chance against HeFiT. In all other SFiT experiments, we have employed hyperparameters Hyp Set 2. The hyperparameters chosen for SFiT are shown in Table~\ref{tab:hyperparameter-configs}. 

\label{app:hyp_space}
\begin{table}[tpb]

\begin{tabular}{|c|c|c|}
\hline
               & Epochs & Hidden Dropout P \\ \hline
SFiT Hyp Set 1 & 4      & 0.10             \\ \hline
SFiT Hyp Set 2 & 10     & 0                \\ \hline
\end{tabular}
\caption{Chosen hyperparameter configurations after grid search}
\label{tab:hyperparameter-configs}

\end{table}

\section{Pseudo Annotation Evaluation}
\label{app:appendix_pseudoannotation}
\begin{table*}[btp]
\resizebox{\linewidth}{!}{\scriptsize
\begin{tabular}{|c|c|c|c|c|c|c|}
\hline
\textbf{Model} & \textbf{$\notin$EN-OV/EN} & \textbf{$\notin$ES-OV/ES} & \textbf{$\notin$IT-OV/IT} & \textbf{$\notin$PT-OV/PT} & \textbf{$\notin$FR-OV/FR} &  \textbf{$\notin$ZH-OV/ZH} \\ \hline
XLM-T SFiT & 0.012/0.017 & 0.002/-0.002                   & 0.000/-0.003                  & 0.000/-0.03                    & -0.002/-0.009              & 0.000/0.011                                                             \\ 
XLM-T HeFiT & 0.013/0.024 & -0.002/-0.006                   & -0.004/-0.015                 & -0.002/0.000                    & 0.000/0.001              & -0.004/0.007                                                             \\
XLM-R SFiT & 0.014/0.021 & -0.002/0.000                   & 0.000/-0.002                & -0.001/-0.004                     & -0.002/-0.003              & -0.001/-0.004                                                      \\ 
XLM-R HeFiT & 0.007/0.015 & -0.004/-0.011                   & 0.000/-0.011                 & -0.002/-0.001                     & -0.002/0.003              & -0.003/-0.003                                                      \\ \hline
\end{tabular}
}
\caption{The columns show differences in performance between fine-tuning models using distant pseudo-annotated (50) and random (50) synthetic tweets in an experiment similar to Figure~\ref{fig:radar_few}. The results denote the overall and omitted language performance in the synthetic few shot scenario evaluated on our internal validation set. Each column shows the results of a different omitted language.}
\label{tab:annotation}
\end{table*}

\section{Final Prediction Model Summaries}

\label{app:appendix_submission}

\begin{table*}[btp]
\resizebox{\linewidth}{!}{\scriptsize
\begin{tabular}{|c|c|c|c|c|c|c|c|c|c|}
\hline
\textbf{ID} & \textbf{Category} & \textbf{ES Data} & \textbf{IT Data} & \textbf{\{EN,ZH,PT,FR\}} & \textbf{Synth NL} & \textbf{Synth HI} & \textbf{Synth KO} & \textbf{Synth AR} &  \textbf{In Submission} \\ \hline
1 & SFewS & 50\%                   & 100\%                  & 100\%                    & {YES}              & {YES}              & {YES}               & {YES}               & HeFiT                                               \\ \hline
2 & SFewS& 50\%                   & 100\%                  & 100\%                     & {}                 & {YES}              & {YES}               & {YES}               & HeFiT                                               \\ \hline
3 & SFewS& 50\%                   & 100\%                  & 100\%                     & {YES}              & {YES}              & {YES}               & {}                  & HeFiT, SFiT                                         \\ \hline
4 & S\{KO, HI\} & 50\%                   & 100\%                  &  100\%                    & {}                 & {YES}              & {YES}               & {}                  & HeFiT, SFiT                                         \\ \hline
5 & ZeroS & 50\%                   & 100\%                  & 100\%                     & {}                 & {}                 & {}                  & {}                  & SFiT                                                \\ \hline
6 & SFewS& 100\%                  & 100\%                  & 100\%                     & {YES}              & {YES}              & {YES}               & {YES}               &                                                     \\ \hline
7 & SFewS& 100\%                  & 100\%                  & 100\%                     & {}                 & {YES}              & {YES}               & {YES}               &                                               \\ \hline
8 & SFewS& 100\%                  & 100\%                  & 100\%                     & {YES}              & {YES}              & {YES}               & {}                  &                                               \\ \hline
9 & S\{KO, HI\} & 100\%                  & 100\%                  & 100\%                     & {}                 & {YES}              & {YES}               & {}                  &                                                     \\ \hline
10 & ZeroS & 100\%                  & 100\%                  & 100\%                     & {}                 & {}                 & {}                  & {}                  & SFiT                                                \\ \hline
11 & SFewS& 100\%                  & 50\%                   & 100\%                     & {YES}              & {YES}              & {YES}               & {YES}               & HeFiT                                                    \\ \hline
12 & SFewS& 100\%                  & 50\%                   & 100\%                     & {}                 & {YES}              & {YES}               & {YES}               & HeFiT                                                    \\ \hline
13 & SFewS& 100\%                  & 50\%                   & 100\%                     & {YES}              & {YES}              & {YES}               & {}                  &                                                     \\ \hline
14 & S\{KO, HI\} & 100\%                  & 50\%                   & 100\%                     & {}                 & {YES}              & {YES}               & {}                  &                                                     \\ \hline
15 & ZeroS& 100\%                  & 50\%                   & 100\%                       & {}                 & {}                 & {}                  & {}                  &                                                     \\ \hline

\end{tabular}
}
\caption{The data compilation of the fifteen model variations of our post-submission evaluations,  repeated for both fine-tuning techniques and pre-trained models. The "In Submission" column indicates which XLM-T fine-tuning procedure was used (if any) for our submission ensemble to the Shared Task}
\label{tab:post_sub_models}
\end{table*}

\end{document}